\documentclass[11pt]{article}

\usepackage[preprint]{acl}

\usepackage{times}
\usepackage{latexsym}

\usepackage[T1]{fontenc}

\usepackage[utf8]{inputenc}

\usepackage{microtype}

\usepackage{inconsolata}

\usepackage{graphicx}
\usepackage{array}
\usepackage{booktabs}
\usepackage{multirow}
\usepackage{makecell}
\usepackage{flushend}
\title{Beyond Averages: Evaluating LLMs on Human Survey Replication at the Distributional Level}

\author{
Jeonghyeon Moon,
Jiwon Kim,
Yeheum Lah,
Yoonju Han,
Yuncheol Kang\thanks{Corresponding author.} \\
Ewha Womans University \\
Seoul, Republic of Korea \\
\texttt{moonjeonghyeon@ewha.ac.kr} \\
\texttt{kim.jiwon@ewha.ac.kr, yhlah@ewha.ac.kr} \\
\texttt{yoonju.han@ewha.ac.kr, yckang@ewha.ac.kr}
}

\begin{document}
\maketitle
\begin{abstract}
LLMs are increasingly used to simulate human survey responses, but prior work has mainly evaluated replication using mean-level or aggregate agreement, offering limited insight into whether LLMs reproduce the variability of human behavior. We evaluate LLM-based survey replication at the distributional level using a non-public 2010 consumer choice experiment on Korean instant noodle purchases, a setting unlikely to overlap with model training data. We evaluate three response variables of differing statistical type: binary purchase incidence, categorical brand choice, and count purchase quantity. For each, we compare human and LLM responses at mean-level, pattern, and distributional alignment, and against reference baselines from the human data alone. LLMs reproduce condition-level patterns reasonably well but fail to capture distributional structure: for purchase quantity, no model beats a condition-insensitive baseline that simply matches the pooled human distribution. Because models that match human means well can still produce distributions further from humans than this baseline, mean-based evaluation alone can be actively misleading. Replication also varies with input configuration, with structured personas and multimodal inputs improving alignment while explicit reasoning prompting degrades it monotonically.
\end{abstract}

\section{Introduction}
Survey-based data collection has long served as a primary tool in the social sciences and marketing for studying human decision-making \citep{krosnick1999}. While surveys enable researchers to directly observe human choices and judgments, they require substantial cost and time for instrument design, response collection, and repeated experimentation\citep{goyder1986, coughlan2009}. These limitations become particularly restrictive when comparing behavioral patterns across many experimental conditions\citep{ebert2018, al-ubaydli2017}. To address these challenges, recent studies have increasingly explored large language models (LLMs) as tools for simulating human responses\citep{lu2025, hewitt2024, horton2023}. Reported applications include predicting social science experiment outcomes, modeling economic decision-making, and generating survey responses\citep{suri2024, aher2023, wang2023}.

However, prior work has mainly evaluated LLM responses through mean tendencies or aggregate group patterns \citep{hewitt2024, luo2024, liusie2024}, offering limited insight into whether LLMs reproduce the variability and heterogeneity of human responses \citep{elangovan2024, wang2025}, which cannot be captured by group means alone \citep{allenby1998, louviere2001}. Recent studies have begun examining distribution-level replication \citep{tjuatja2024, bisbee2024, qin2026}, but typically focus on a single response type and rely on public survey datasets, leaving it unclear whether findings generalize across response types of differing statistical properties or to data outside likely training distributions.

To address these gaps, we evaluate LLM-based survey replication using a non-public 2010 consumer choice experiment involving Korean instant noodle purchases under 12 promotional conditions. The experiment includes three response variables with different statistical properties: purchase incidence (binary), brand choice (categorical), and purchase quantity (count). We repeatedly sample LLM responses under matched conditions and compare the resulting distributions with human responses.

We evaluate replication at three levels of alignment for each response type: mean-level alignment, condition-level pattern alignment, and distributional alignment. Using type-appropriate metrics, we compare human and LLM responses across binary, categorical, and count outcomes. We further examine how replication performance varies across input configurations, including input modality, persona format, and reasoning prompting.

The contributions of this paper are threefold. First, we propose a multi-type, multi-level evaluation framework for LLM-based survey replication spanning binary, categorical, and count response variables, and use it to show that LLMs reproduce condition-level preference patterns reasonably well but fail to capture the distributional structure of human responses, particularly the variability and tail behavior of purchase quantity. Second, we provide empirical evidence on the capabilities and limitations of LLMs using a non-public behavioral dataset that reduces concerns about training-data contamination. Third, through a factorial analysis of input configuration, we identify a consistent pattern, observed across model families and scales, in which explicit reasoning prompting reduces distributional alignment, offering practical implications for prompt design in LLM-based behavioral simulation.

\begin{figure*}[t]
\centering
\includegraphics[width=\textwidth]{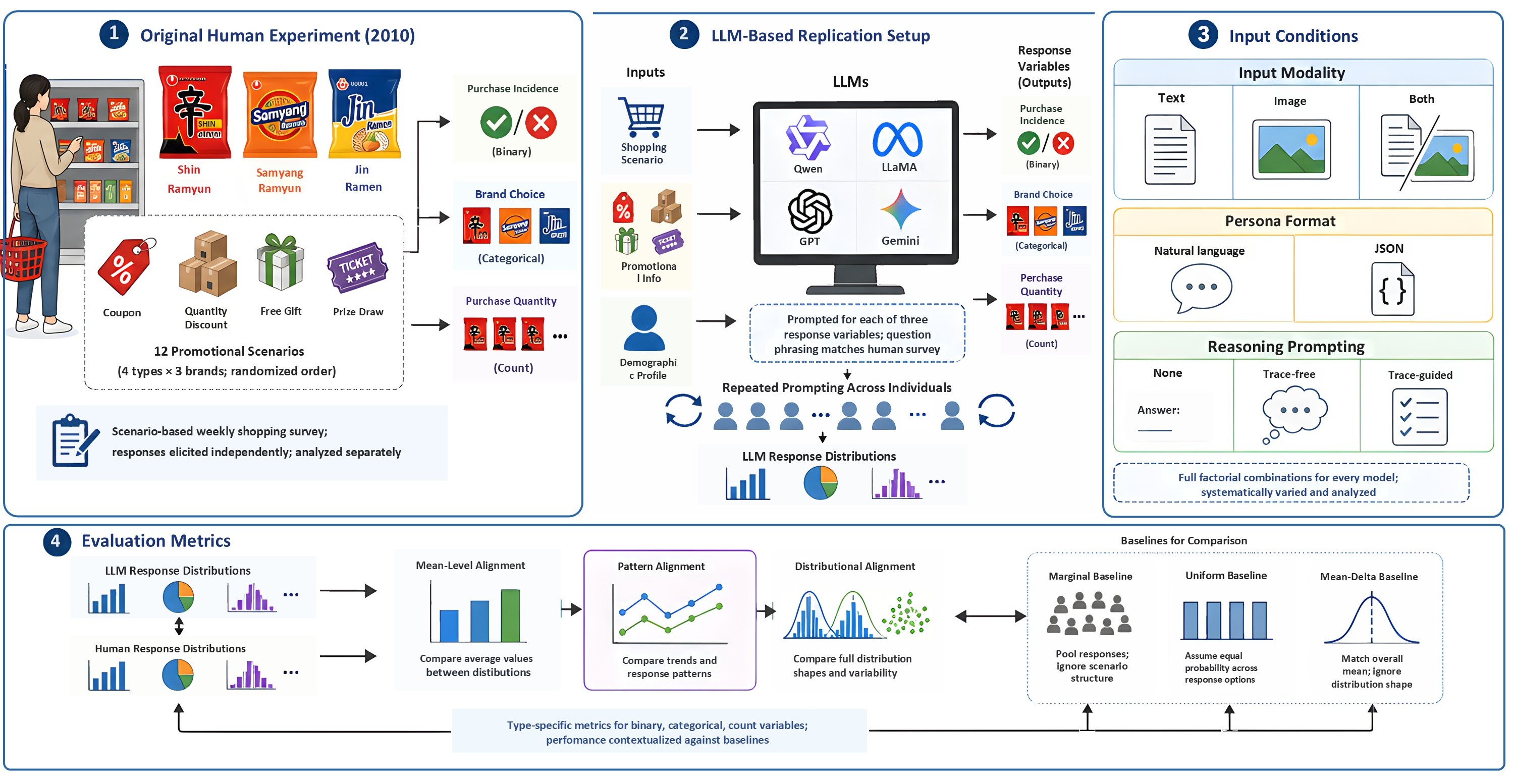}
\caption{
Overview of the human consumer choice experiment and the LLM-based survey replication framework. 
(1) Human responses were collected under 12 promotional conditions across three response variables: purchase incidence, brand choice, and purchase quantity. 
(2) LLM-based replication was conducted by repeatedly prompting multiple models using shopping scenarios, promotional information, and demographic profiles to generate response distributions. 
(3) Input configurations systematically varied across input modality, persona format, and reasoning prompting conditions. 
(4) LLM-generated and human response distributions were compared using mean-level, pattern, and distributional alignment metrics, alongside three reference baselines (marginal, uniform, and mean-delta).
}
\label{fig:overview}
\end{figure*}

\section{Related Work}

Recent research has examined whether LLMs can simulate human behavior across a range of domains, from classic economic decision-making experiments\citep{horton2023} to social science replications that approximate group-level treatment effects\citep{hewitt2024, aher2023}, with further efforts to improve fidelity through fine-tuning or reasoning traces\citep{lu2025}. A related line of work uses LLMs as survey respondents conditioned on demographic, ideological, or cultural attributes to approximate population-level opinions and choices\citep{argyle2023, hwang2023, kwok2024, zhao2024}.

However, some limitations remain in the evaluation methodology used in this literature. First, most studies assess alignment primarily at the level of mean responses or aggregate group patterns\citep{liusie2024, hewitt2024}, which fails to capture the variability and probabilistic structure of human responses. A small number of recent studies move beyond mean-level evaluation to the distributional level\citep{tjuatja2024, bisbee2024, qin2026}, but typically focus on a single response type such as a Likert-scale item, leaving open whether the same patterns hold for behavioral tasks involving response variables of differing statistical character. Second, much of the prior work relies on publicly available surveys that may overlap with LLM training corpora, making it difficult to separate genuine simulation from memorization\citep{wang2025, bisbee2024}. Third, while simple prompt designs are known to constrain replication quality\citep{qu2024, bisbee2024}, recent prompt features such as multimodal input and explicit reasoning prompting have not been systematically evaluated for their effect on distributional alignment.

This study addresses these gaps by evaluating LLM replication across three response variables of differing statistical type (i.e., binary, categorical, and count) using type-appropriate metrics at mean-level, pattern, and distributional levels; by using a non-public 2010 consumer choice experiment as the reference dataset to mitigate contamination concerns; and by conducting a factorial analysis of input configuration that reveals a previously under-reported pattern in which explicit reasoning traces can degrade distributional alignment.

\section{Experimental Setup}
\subsection{Original Human Experiment}
\label{sec:human_experiment}
The human data used in this study come from a consumer choice experiment conducted in 2010 to examine purchase behavior under promotional conditions. Because the dataset has not been publicly released, it is unlikely to have appeared in the training data of the evaluated LLMs, reducing concerns about data contamination.

The experiment was structured as a scenario-based weekly grocery shopping survey involving three Korean instant noodle brands (Shin Ramyun, Samyang Ramen, and Jin Ramen). Respondents were exposed to 12 promotional conditions constructed from four promotion types (coupon, quantity discount, free gift, and prize draw) across the three brands.
For each condition, three response variables were collected independently: (1) purchase incidence, indicating whether the respondent would purchase any brand; (2) brand choice, indicating which brand would be selected; and (3) purchase quantity, indicating how many units would be purchased. The three variables differ in statistical type, corresponding to binary, categorical, and count responses. Additional details on respondent demographics and the survey instrument are provided in Appendix~\ref{app:human_participants}.

\subsection{LLM-based Replication Setup}
To evaluate whether LLMs can replicate the human responses described in Section~\ref{sec:human_experiment}, we design a replication procedure in which LLMs are prompted to perform the same response task under identical conditions. For each promotional condition, the model receives a prompt containing the shopping scenario, the promotional information, and a set of demographic attributes representing a respondent profile, and generates a response for each of the three response variables.
Consistent with the structure of the original survey, the three response variables are elicited as independent items rather than as a chained decision. Specifically, for each response variable, the model is prompted separately with a question phrased to match the corresponding survey item, so that the generation of one response does not condition on another.
To enable distributional comparison rather than point-estimate evaluation, each prompt configuration is executed repeatedly across a set of demographic profiles drawn to reflect the composition of the human respondent pool. The resulting samples are aggregated to construct a response distribution for each model × input condition × promotional condition combination, which is then compared with the corresponding human response distribution. All models are evaluated under a consistent procedure, while variations in input format and prompt content are analyzed separately as described in Section~\ref{sec:experimental_input_conditions}.

\subsection{Input Conditions}
\label{sec:experimental_input_conditions}
To examine how input configuration affects LLM-generated responses, we vary prompts along three dimensions while holding the underlying task and promotional conditions fixed.
The first dimension is input modality. Promotional information is presented as text only, image only, or image+text. The second dimension is persona format, where respondent attributes are provided either as a natural-language description or as a structured JSON object. The third dimension is reasoning prompting. We compare three settings: (a) None, in which the model generates only the response value; (b) trace-free, in which the model generates an unconstrained reasoning trace before answering; and (c) trace-guided, in which the model generates a structured reasoning trace before producing the final response.
The full factorial combination of these dimensions is applied to every model, and results are analyzed both in aggregate and by condition.

\begin{table}[!t]
\centering
\footnotesize
\setlength{\tabcolsep}{3pt}

\begin{tabular}{
>{\raggedright\arraybackslash}m{1.7cm}
>{\raggedright\arraybackslash}m{1.8cm}
>{\raggedright\arraybackslash}m{1.8cm}
>{\raggedright\arraybackslash}m{1.8cm}
}
\toprule

\textbf{Response Type}
& \textbf{Mean-level}
& \textbf{Pattern}
& \textbf{Distributional} \\

\midrule

Purchase Incidence (binary)
& Absolute error of purchase rate
& Pearson $r$ on purchase rates across conditions
& JSD on Bernoulli distribution \\

\midrule

Brand choice (categorical)
& Mean absolute error on per-brand selection rates
& Pearson $r$ on brand-condition selection patterns
& JSD on brand choice distributions \\

\midrule

Quantity (count)
& MAE on mean quantity
& Pearson $r$ on mean quantity across conditions
& Wasserstein distance \\

\bottomrule
\end{tabular}

\caption{
Evaluation metrics used for each response type across the three alignment levels.
}

\label{tab:metrics}
\end{table}

\subsection{Models}
We evaluate open- and closed-source LLMs spanning the Qwen, LLaMA, GPT, and Gemini families, all supporting multimodal input. All models were evaluated under the same prompting and evaluation protocol; full details are in Appendix~\ref{app:model_details}.

\subsection{Evaluation Metrics}
\label{sec:evaluation_metrics}
Because the three response variables differ in statistical type, we apply type-appropriate metrics rather than a single uniform metric. For each variable, we evaluate replication performance at three levels of alignment: mean-level, pattern, and distributional.

Mean-level alignment captures whether the LLM reproduces the average response under each condition. Pattern alignment captures whether LLM and human responses exhibit consistent patterns of relative magnitude across the 12 promotional conditions. We use the term pattern alignment rather than directional alignment because the 12 conditions are factorial combinations of promotion type and brand and do not lie on a single ordered dimension; the metric therefore measures the consistency of condition-level response patterns rather than agreement on an underlying direction. Distributional alignment captures whether the LLM reproduces the full shape of the response distribution under each condition, including variability and heterogeneity around the mean. For distributional metrics, we additionally compute 95\% bootstrap confidence intervals by resampling the 12 promotional conditions with replacement 10,000 times. Bootstrap confidence intervals are reported in Appendix~\ref{app:bootstrap_ci}. The metrics applied to each response type are summarized in Table \ref{tab:metrics}. Distributional metrics are computed for every model × input condition × promotional condition cell, and then averaged across promotional conditions for model-level and condition-level comparison.

\begin{table}[!t]
\centering
\footnotesize
\setlength{\tabcolsep}{2pt}
\renewcommand{\arraystretch}{1.2}

\begin{tabular}{
>{\raggedright\arraybackslash}m{2.2cm}
>{\centering\arraybackslash}m{1.5cm}
>{\centering\arraybackslash}m{1.7cm}
>{\centering\arraybackslash}m{1.7cm}
}
\hline
\multirow{2}{*}{\centering\textbf{Model}}
& \textbf{Mean-level}
& \textbf{Pattern}
& \textbf{Distributional} \\
& \textbf{(MAE)}
& \textbf{(Pearson $r$)}
& \textbf{(Mean JSD)} \\
\hline

Gemini-2.5-Pro & 0.023 & 0.587* & 0.003 \\
Qwen3-VL-30B-A3B-Instruct & 0.039 & -0.274 & 0.013 \\
Qwen2.5-VL-7B-Instruct & 0.055 & 0.029 & 0.018 \\
GPT-5-nano & 0.046 & 0.150 & 0.021 \\
Qwen2.5-VL-72B-Instruct & 0.043 & -0.291 & 0.021 \\
GPT-5.2 & 0.048 & -0.197 & 0.024 \\
Llama-3.2-11B-Vision-Instruct & 0.048 & -- & 0.025 \\
Qwen3-VL-8B-Instruct & 0.048 & -- & 0.025 \\
GPT-4.1-nano & 0.048 & -- & 0.025 \\
Llama3-LLaVA-Next-8B-HF & 0.048 & -- & 0.025 \\

\hline
Marginal baseline & 0.013 & 0.000 & 0.001 \\
Mean-delta baseline & 0.000 & 1.000 & 0.000 \\
Uniform baseline & 0.452 & 0.000 & 0.209 \\
\hline
\end{tabular}

\caption{
Binary purchase incidence replication performance by model and baseline.
Pattern alignment is measured using Pearson correlation across the 12 promotional conditions.
Significance markers indicate correlation tests (* $p < 0.05$).
}

\label{tab:binary_incidence}
\end{table}

To contextualize the magnitude of LLM-human distances and to provide reference points across the three alignment levels, we report three baselines computed from the human data alone. The marginal baseline predicts, under every condition, the overall human response distribution pooled across all conditions. It represents a condition-insensitive predictor that preserves the shape of the human distribution but ignores condition-specific variation. The uniform baseline predicts a uniform distribution over the response support under every condition, representing an uninformed predictor that uses no information about human responses. The mean-delta baseline is condition-specific: for each condition, it places all probability mass at the human mean response of that condition. For categorical brand choice, where no natural mean exists, the baseline instead assigns all probability mass to the modal human-selected brand. This represents a degenerate predictor that matches the condition-level central tendency exactly but assumes zero response variability.

These three baselines differ systematically in the information they use, and consequently in their performance across the three alignment levels. By construction, mean-delta achieves zero mean-level error and pattern correlation of 1.0 but produces sharp distributions; marginal and uniform yield zero pattern correlation, but only marginal preserves the human distribution shape. Comparing against these baselines lets us assess whether LLM responses are condition-sensitive (vs. marginal), informative (vs. uniform), and variability-preserving (vs. mean-delta).

\section{Results}
Throughout this section, we report LLM performance alongside the three reference baselines (marginal, uniform, mean-delta), which serve as interpretive anchors rather than competing predictors.

\subsection{Replication Performance by Response Type}
\subsubsection{Purchase Incidence (Binary)}
Purchase incidence captures whether respondents would buy any of the three brands under each promotional condition. We evaluate mean absolute error of condition-level purchase rates (mean-level), Pearson correlation of purchase rates across the 12 conditions (pattern), and JSD on Bernoulli distributions (distributional). Table \ref{tab:binary_incidence} reports the results.
Two findings stand out. First, the marginal baseline achieves a lower MAE (0.013) than every evaluated model, indicating that no LLM improves on a condition-insensitive prediction of the pooled human purchase rate. Although several models reach low absolute errors (Gemini-2.5-Pro 0.023, Qwen3-VL-30B-A3B-Instruct 0.039), none captures condition-level variation well enough to beat simply predicting the overall human rate.
Second, pattern alignment is weak across models. Only Gemini-2.5-Pro shows a statistically significant positive correlation (r = 0.587, p = 0.045). Three models invert the human pattern, producing negative correlations (Qwen2.5-VL-72B-Instruct -0.291, Qwen3-VL-30B-A3B-Instruct -0.274, GPT-5.2 -0.197), and four models predict purchase under every condition, yielding zero variance and undefined correlations. This last behavior, present across both small and large models, reflects a systematic tendency to over-predict purchase that erases condition-level variation. Distributional differences are correspondingly small in absolute terms (most models between 0.01 and 0.02 JSD), but this reflects the limited range of a binary variable rather than faithful replication, since the uniform baseline reaches 0.209 while the marginal baseline reaches only 0.001.
Taken together, these results indicate that current LLMs struggle to reproduce condition-level variation in purchase incidence: the majority either invert the human pattern or collapse to a constant positive response, and none outperforms a condition-insensitive baseline.

\begin{table}[!t]
\centering
\footnotesize
\setlength{\tabcolsep}{2pt}
\renewcommand{\arraystretch}{1.2}

\begin{tabular}{
>{\raggedright\arraybackslash}m{2.2cm}
>{\centering\arraybackslash}m{1.5cm}
>{\centering\arraybackslash}m{1.7cm}
>{\centering\arraybackslash}m{1.7cm}
}
\hline
\multirow{2}{*}{\centering\textbf{Model}}
& \textbf{Mean-level}
& \textbf{Pattern}
& \textbf{Distributional} \\
& \textbf{(MAE)}
& \textbf{(Pearson $r$)}
& \textbf{(Mean JSD)} \\
\hline

Qwen2.5-VL-72B-Instruct & 0.055 & 0.957 & 0.014 \\
Llama-3.2-11B-Vision-Instruct & 0.107 & 0.849 & 0.030 \\
Qwen3-VL-8B-Instruct & 0.103 & 0.863 & 0.031 \\
Qwen3-VL-30B-A3B-Instruct & 0.122 & 0.828 & 0.040 \\
GPT-5-nano & 0.121 & 0.877 & 0.056 \\
GPT-5.2 & 0.133 & 0.835 & 0.075 \\
Gemini-2.5-Pro & 0.160 & 0.868 & 0.081 \\
GPT-4.1-nano & 0.190 & 0.731 & 0.103 \\
Llama3-LLaVA-Next-8B-HF & 0.296 & 0.580 & 0.262 \\
Qwen2.5-VL-7B-Instruct & 0.364 & 0.029 & 0.363 \\

\hline
Marginal baseline & 0.180 & 0.000 & 0.074 \\
Mean-delta baseline & 0.242 & 1.000 & 0.216 \\
Uniform baseline & 0.211 & 0.000 & 0.109 \\
\hline
\end{tabular}

\caption{
Brand choice replication performance by model and baseline.
All reported Pearson correlations except Qwen2.5-VL-7B-Instruct are significant at $p < 0.001$.
}

\label{tab:brand_choice}
\end{table}
\subsubsection{Brand Choice (Categorical)}
Brand choice captures which of the three brands was selected under each promotional condition. We evaluate mean absolute error of per-brand selection rates (mean-level), Pearson correlation of brand-condition selection patterns (pattern), and JSD on the three-brand choice distribution (distributional). Table \ref{tab:brand_choice} reports the results.

Brand choice is the response type on which LLMs perform most strongly. Several models substantially outperform all three baselines on both mean-level and distributional metrics. Qwen2.5-VL-72B-Instruct achieves the lowest MAE (0.055), more than three times lower than the marginal baseline (0.180), and the lowest JSD (0.014), well below the marginal baseline (0.074). Pattern correlations exceed 0.85 for several models, with Qwen2.5-VL-72B-Instruct reaching 0.957. This indicates that LLMs effectively differentiate brand selection across promotional conditions, possibly because brand-promotion associations are explicit in the prompt and align with common consumer behavior patterns reflected in the training data. At the lower end, Qwen2.5-VL-7B-Instruct shows almost no pattern alignment (0.029) and the highest JSD (0.363).
All models except Qwen2.5-VL-7B-Instruct showed statistically
significant positive pattern correlations (all p < 0.001).
The baseline behavior on this task is itself informative. The mean-delta baseline, which places all probability mass on the modal brand for each condition, produces both a higher MAE (0.242 vs 0.180) and a higher JSD (0.216 vs 0.074) than the marginal baseline. While mean-delta captures the modal brand perfectly, it discards the proportions of non-modal brands, which inflates both metrics. This shows that even a predictor with perfect modal alignment can be outperformed by a condition-insensitive predictor when the full categorical distribution is considered, illustrating the gap between mode-level and distributional evaluation. Unlike incidence and quantity, however, the strongest LLMs surpass all baselines here, indicating genuine condition-sensitive replication of categorical choice while preserving reasonable distributional structure over the three-brand support.

\subsubsection{Purchase Quantity (Count)}

\begin{table}[!b]
\centering
\footnotesize
\setlength{\tabcolsep}{2pt}
\renewcommand{\arraystretch}{1.2}

\begin{tabular}{
>{\raggedright\arraybackslash}m{2.2cm}
>{\centering\arraybackslash}m{1.5cm}
>{\centering\arraybackslash}m{1.5cm}
>{\centering\arraybackslash}m{2.0cm}
}
\hline
\textbf{Model}
& \raisebox{-4pt}{\shortstack{\textbf{Mean-level}\\\textbf{(MAE)}}}
& \raisebox{-4pt}{\shortstack{\textbf{Pattern}\\\textbf{(Pearson $r$)}}}
& \raisebox{-3pt}{\shortstack{\textbf{Distributional}\\\textbf{(Mean}\\[-1pt]\textbf{Wasserstein)}}} \\
\hline

Gemini-2.5-Pro & 1.425 & 0.768** & 1.467 \\
Llama-3.2-11B-Vision-Instruct & 2.098 & 0.121 & 1.669 \\
Qwen3-VL-30B-A3B-Instruct & 1.672 & 0.584* & 1.751 \\
Qwen3-VL-8B-Instruct & 0.726 & 0.540 & 1.810 \\
GPT-5-nano & 1.673 & 0.125 & 1.819 \\
GPT-5.2 & 1.901 & 0.477 & 1.968 \\
Qwen2.5-VL-72B-Instruct & 1.993 & 0.416 & 2.064 \\
Qwen2.5-VL-7B-Instruct & 2.535 & 0.333 & 2.503 \\
GPT-4.1-nano & 2.631 & -0.405 & 2.977 \\
Llama3-LLaVA-Next-8B-HF & 2.945 & 0.076 & 3.155 \\

\hline
Marginal baseline & 0.582 & 0.000 & 0.641 \\
Mean-delta baseline & 0.000 & 1.000 & 2.665 \\
Uniform baseline & 4.481 & 0.000 & 4.623 \\
\hline
\end{tabular}

\caption{
Purchase quantity replication performance by model and baseline.
Significance markers indicate Pearson correlation tests
(* $p < 0.05$, ** $p < 0.01$).
}

\label{tab:purchase_quantity}
\end{table}

\begin{figure*}[t]
\centering
\includegraphics[width=0.9\textwidth]{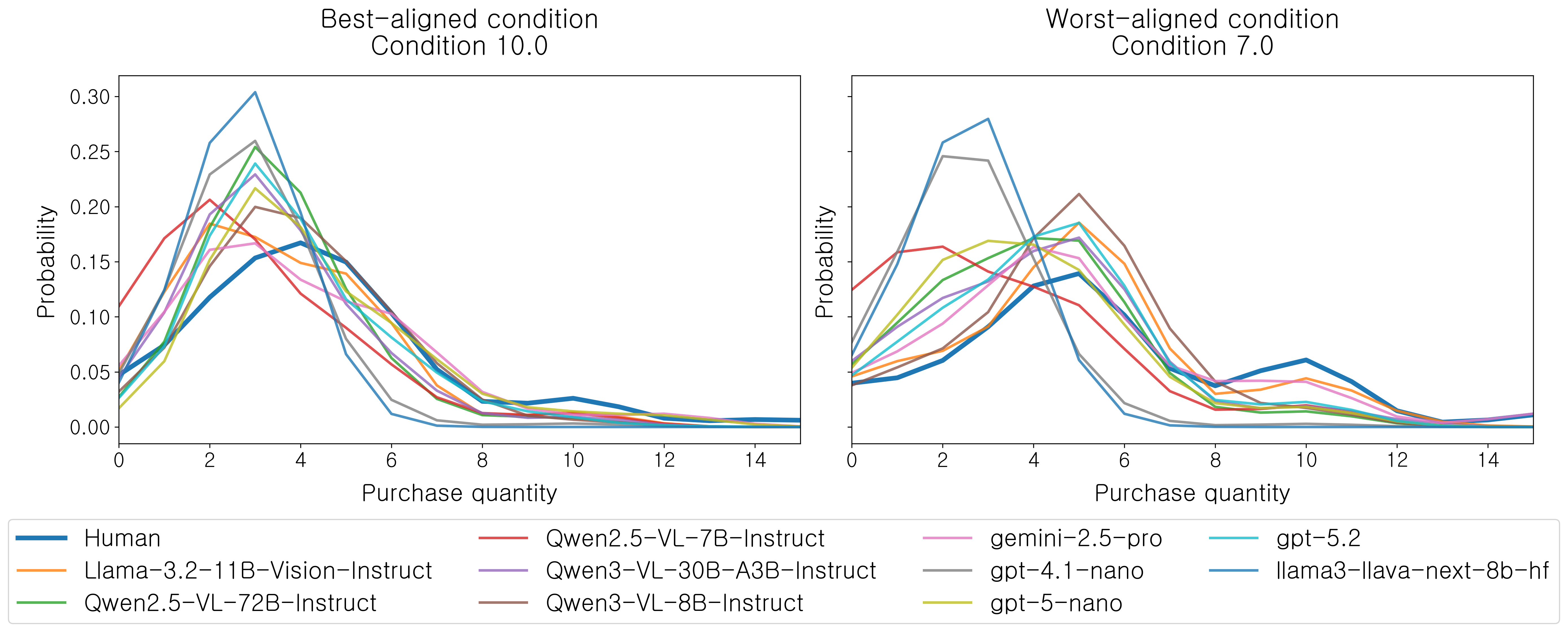}
\caption{
Example purchase quantity distributions for the best-aligned and worst-aligned promotional conditions based on Wasserstein distance. 
The left panel shows a condition with relatively strong agreement between human and LLM-generated quantity distributions, while the right panel illustrates a poorly aligned condition with substantial differences in peak location and distributional spread.
}
\label{fig:quantity_histogram}
\end{figure*}

Purchase quantity captures how many units were purchased under each promotional condition. We evaluate MAE on mean quantity (mean-level), Pearson correlation of mean quantities across conditions (pattern), and Wasserstein distance on quantity distributions (distributional). Table \ref{tab:purchase_quantity} reports the results, and Figure \ref{fig:quantity_histogram} illustrates the best-aligned and worst-aligned conditions.
Several models approximate human mean quantities reasonably well, with Qwen3-VL-8B-Instruct achieving the lowest MAE (0.726) and Gemini-2.5-Pro the highest pattern correlation (0.768). Yet the central finding for this response type concerns distributional alignment. The marginal baseline, which predicts the pooled human distribution under every condition without using any condition-specific information, achieves a mean Wasserstein distance of 0.641, lower than every evaluated model. The best-performing model (Gemini-2.5-Pro, 1.467) is more than twice as far from the human distributions. Despite incorporating condition-specific prompts, LLM-generated quantity distributions are further from human distributions than a condition-insensitive predictor that simply matches the overall human distribution.
This pattern indicates that LLMs introduce condition sensitivity at the cost of distributional fidelity: their condition-specific mean adjustments are accompanied by a collapse of distributional spread that more than offsets any gain from mean matching. The position of the models between the marginal (0.641) and mean-delta (2.665) baselines makes this concrete. All models fall between these two reference points, with most closer to mean-delta, indicating that LLM quantity responses behave more like condition-specific point-mass predictors than like shape-preserving predictors. This provides direct quantitative evidence for the central argument of this study: evaluating LLM-based survey replication at the mean level alone can be actively misleading, because models that match human means well may simultaneously produce distributions further from human responses than a condition-insensitive baseline.

\subsubsection{Cross-Response-Type Patterns}
Two patterns hold across response types. First, model rankings are largely consistent: Gemini-2.5-Pro, the Qwen3-VL series, and GPT-5-nano are the strongest performers, while GPT-4.1-nano, llama3-llava-next-8b-hf, and Qwen2.5-VL-7B-Instruct are the weakest. Second, and more importantly, the marginal and mean-delta baselines dominate at opposite alignment levels: mean-delta is best at the mean level by construction, while marginal is best at the distributional level for both incidence and quantity. Because the choice of metric can thus reverse the apparent ranking of predictors, evaluating LLM replication at a single alignment level is insufficient. This asymmetry motivates the multi-level evaluation framework adopted here.

\subsection{Effect of Input Conditions}
\label{sec:input_conditions}
To analyze how input configuration affects replication, we compare mean Pearson correlation and mean JSD across the three input dimensions defined in Section~\ref{sec:experimental_input_conditions}: input modality, persona format, and reasoning prompting. For each dimension, we aggregate across all models, promotional conditions, and response types, providing a model-agnostic view of how input configuration shifts replication fidelity. Tables \ref{tab:persona_format} and \ref{tab:reasoning_type} report the persona format and reasoning results; input modality results are reported in Appendix~\ref{app:modality_results}. The three dimensions differ markedly in the size of their effects. Input modality has the smallest effect, with image-based conditions marginally outperforming text-only on both metrics (a gap of about 0.04 in Pearson and 0.01 in JSD; Appendix~\ref{app:modality_results}). Persona format has a larger effect: structured JSON personas outperform free-text personas on both metrics (0.800 vs 0.720 Pearson, 0.237 vs 0.283 JSD), improving pattern correlation by 0.08 and reducing JSD by 0.046. Because the underlying demographic information is identical across the two formats, this indicates that the structural representation of persona input, not its content, affects how LLMs use demographic information. Structured representations may make individual attributes more salient and reduce parsing overhead, leading to more consistent condition-specific responses. Reasoning prompting produces the most notable pattern. Alignment decreases monotonically as the degree of structured reasoning increases: the no-reasoning setting achieves the best alignment (0.759 Pearson, 0.264 JSD), followed by trace-free (0.722, 0.286), and trace-guided performs worst (0.692, 0.290). The gap between no-reasoning and trace-guided is the largest single effect across all three dimensions (0.067 in Pearson and 0.026 in JSD). This is contrary to the common assumption that explicit reasoning improves performance. A plausible interpretation is that reasoning traces commit the model to a specific justification, narrowing the space of plausible final answers and collapsing the response distribution. The fact that the more rigid trace-guided condition is worse than the unstructured trace-free condition reinforces this reading: greater reasoning structure leads to greater distributional collapse. Taken together, these results indicate that replication fidelity is sensitive to design choices beyond model selection, with structural and prompt-design factors such as persona format and reasoning mattering more than input modality. All three dimensions move pattern and distributional alignment in the same direction, and the reasoning effect in particular has a practical implication: when the goal is to reproduce response variability across hypothetical respondents rather than to produce a single best answer, prompts that elicit direct responses without externalized deliberation are preferable to reasoning-augmented prompts.

\begin{table}[!t]
\centering
\footnotesize
\setlength{\tabcolsep}{3pt}
\renewcommand{\arraystretch}{1.2}

\begin{tabular}{
>{\raggedright\arraybackslash}m{2.1cm}
>{\centering\arraybackslash}m{1.7cm}
>{\centering\arraybackslash}m{1.7cm}
}
\hline
\centering\textbf{Persona format}
& \textbf{Mean Pearson $r$}
& \textbf{Mean JSD} \\
\hline

JSON & 0.800 & 0.237 \\
Text & 0.720 & 0.283 \\

\hline
\end{tabular}

\caption{
Average pattern and distributional alignment across persona formats.
}

\label{tab:persona_format}
\end{table}

\begin{table}[!t]
\centering
\footnotesize
\setlength{\tabcolsep}{3pt}
\renewcommand{\arraystretch}{1.2}

\begin{tabular}{
>{\raggedright\arraybackslash}m{2.1cm}
>{\centering\arraybackslash}m{1.7cm}
>{\centering\arraybackslash}m{1.7cm}
}
\hline
\centering\textbf{Reasoning type}
& \textbf{Mean Pearson $r$}
& \textbf{Mean JSD} \\
\hline

None & 0.759 & 0.264 \\
Trace-free & 0.722 & 0.286 \\
Trace-guided & 0.692 & 0.290 \\

\hline
\end{tabular}

\caption{
Average pattern and distributional alignment across reasoning prompting conditions.
}

\label{tab:reasoning_type}
\end{table}

\section{Discussion}
\subsection{What LLMs Capture and Fail to Capture}
LLMs reproduce some aspects of human survey responses but systematically miss others. On the capture side, several models achieve pattern correlations above 0.85 on brand choice, indicating that they recover the relative attractiveness of brand-promotion combinations across conditions, and Gemini-2.5-Pro reaches a pattern correlation of 0.77 on quantity. These results show that LLMs can produce condition-sensitive responses in factorial behavioral tasks, particularly for categorical choices where promotion-brand associations align with common patterns reflected in the training data.
On the failure side, three patterns recur across response types. On incidence, most models produce zero-variance or pattern-inverted responses, failing to differentiate purchase likelihood across conditions in a way that aligns with humans. On quantity, no model matches the distributional alignment of the marginal baseline, indicating that the condition sensitivity introduced by prompting comes at the cost of distributional spread that exceeds any gain from condition-level mean adjustment. Both patterns point to a systematic tendency toward mode-seeking generation: LLMs behave more like condition-specific point-mass predictors than like distribution-preserving ones, capturing relative tendencies across conditions while collapsing the variability that characterizes human responses.

\subsection{Implications for Survey Replication}
These results position LLM-based survey replication as a supplementary tool for exploring relative response patterns rather than a full substitute for human surveys. Two implications follow. First, the asymmetry between marginal and mean-delta baselines means that a researcher relying on mean-level metrics alone would judge LLMs to approximate human quantity responses well, even though the same models underperform the marginal baseline distributionally. Distributional evaluation with explicit baseline comparison is therefore necessary to avoid this misdirection. Second, the reasoning effect runs counter to the assumption that explicit reasoning improves performance: reasoning traces appear to commit the model to a single justification, narrowing the response distribution. When the goal is to reproduce response heterogeneity rather than a single best answer, direct-answer prompts are preferable to reasoning-augmented ones.

\section{Conclusion}
This study evaluated how well LLMs replicate human survey responses across three response variables of differing statistical type, comparing LLM and human responses at three alignment levels (mean, pattern, distributional) against three reference baselines. A consistent pattern emerges: LLMs reproduce condition-level response patterns reasonably well but fall short distributionally. For purchase quantity, no model beats a condition-insensitive marginal baseline despite using condition-specific input; for incidence, most models produce zero-variance or pattern-inverted responses; only for brand choice do top models surpass all baselines. Input configuration also matters, with structured personas and multimodal inputs improving alignment while explicit reasoning prompting degrades it monotonically. These findings position LLM-based replication as a tool for exploring relative response patterns rather than a substitute for human surveys, and show that mean-based evaluation alone can be actively misleading, underscoring the need for distributional evaluation with explicit baseline comparison.

\section*{Limitations}
This study has several limitations. First, because the experiments were conducted using a single consumer choice dataset, the findings may not fully generalize to other behavioral domains or survey contexts. Second, the response distributions were constructed by repeatedly sampling outputs from the same prompts, which does not fully capture the long-term context or dynamic interactions involved in real human response processes. Third, the promotional manipulation in our reference dataset produces only modest condition-level variation in human responses, which makes the marginal baseline a particularly strong reference point. In behavioral domains with larger condition effects, the relative ordering of baselines and LLMs may differ. Fourth, our evaluation focuses on aggregate distribution alignment and does not assess whether LLMs reproduce individual-level heterogeneity conditional on demographic attributes. Future work could examine persona-conditional replication using human data stratified by demographic group.

\bibliography{references}

\appendix

\section{Experimental Details} 

\subsection{Human Participant Information}
\label{app:human_participants}
The reference dataset used in this study was derived from a consumer survey conducted by our research group in Korea in 2010. The dataset consisted of 52 respondent profiles containing demographic attributes, baseline brand preference allocations, household consumption information, and attitudinal/personality measures. All respondents completed the full set of 12 promotional conditions, yielding repeated observations across multiple purchase scenarios.

The collected demographic variables included gender, occupation, marital status, age, household size, monthly income, and weekly ramen consumption. The respondent pool consisted of 35 male and 17 female participants. The mean respondent age was 24.9 years (SD = 1.75), with ages ranging from 20 to 29. The survey also collected several measures related to ramen involvement and purchasing tendencies, including interest in ramen products, perceived relevance of ramen to daily life, adventurousness, preference for purchase certainty, and whether the respondent directly purchases food products.

Before the promotional choice tasks, respondents reported baseline preference weights for the three ramen brands (Shin Ramyun, Samyang Ramen, and Jin Ramen), constrained to sum to 100\%. These preference allocations were provided separately in the prompt as respondent-specific baseline preferences. For each respondent, two persona representations were constructed: (1) a natural-language text profile and (2) a structured JSON profile containing the same underlying information. These representations were used as inputs for the persona-format comparison described in Section~\ref{sec:input_conditions}.

The survey used a scenario-based grocery shopping setting in which respondents imagined purchasing one week's worth of groceries at a supermarket. Under each promotional condition, respondents viewed the three ramen products together with one promotional offer applied to a single brand, and then answered the purchase incidence, brand choice, and purchase quantity questions independently.

\subsection{Promotional Conditions}
Table~\ref{tab:promo_conditions} summarizes the promotional conditions used in the survey.

The survey used a factorial design consisting of three ramen brands and four promotional types, resulting in 12 promotional conditions. The three brands were Shin Ramyun, Samyang Ramen, and Jin Ramen. The four promotional types were prize draws, quantity discounts, coupons, and free gifts. Each condition applied a single promotional offer to one brand while the remaining brands were shown without promotion.

The promotional conditions were designed to reflect realistic retail promotions commonly used in Korean grocery markets. Quantity discount promotions used a "buy two, get one free" format, while coupon promotions provided approximately 35\% discounts relative to the original product price. Prize-draw promotions required respondents to submit a lucky number via SMS for entry into a lottery, and free-gift promotions offered one of several household-related premium items. To reduce potential order effects, the presentation order of conditions was varied across respondents.
\begin{table}[t]
\centering
\footnotesize
\setlength{\tabcolsep}{1pt}
\renewcommand{\arraystretch}{1.15}

\begin{tabular}{lccc}
\hline
\textbf{Promotion Type} & \textbf{Shin} & \textbf{Samyang} & \textbf{Jin} \\
\hline

Prize draw 
& SMS lottery 
& SMS lottery 
& SMS lottery \\

Quantity discount 
& 2+1 
& 2+1 
& 2+1 \\

Coupon 
& 35\% off 
& 35\% off 
& 35\% off \\

Free gift 
& Premium item 
& Premium item 
& Premium item \\

\hline
\end{tabular}

\caption{Factorial structure of the 12 promotional conditions.}
\label{tab:promo_conditions}

\end{table}

\subsection{Prompt Templates}
The prompting framework was designed to closely mirror the structure of the original human survey. Each prompt consisted of a system prompt and a user prompt. The system prompt provided respondent information, baseline brand preference allocations, and survey instructions adapted from the original questionnaire. The user prompt presented one promotional condition together with product prices and asked the model to answer the brand-choice and purchase-quantity questions. Details of persona representations are described in Appendix~\ref{app:persona_representation}.

To evaluate the effect of input modality, three prompt configurations were constructed. The text-only condition used textual descriptions of the shopping environment and products. The image condition additionally provided a supermarket ramen-shelf image representing the shopping context, while the image+text condition combined both forms of contextual information. Aside from these modality-specific elements, all prompt components remained identical across conditions. Reasoning-specific instructions are described in Appendix~\ref{app:reasoning_variants}.

Models were instructed to return only two numeric responses corresponding to the selected brand and intended purchase quantity. Example prompts are shown below. Figure~\ref{fig:ramen_shelf} presents the supermarket ramen-shelf image used in image-based conditions.

\begin{figure}[t]
\centering
\includegraphics[width=\columnwidth]{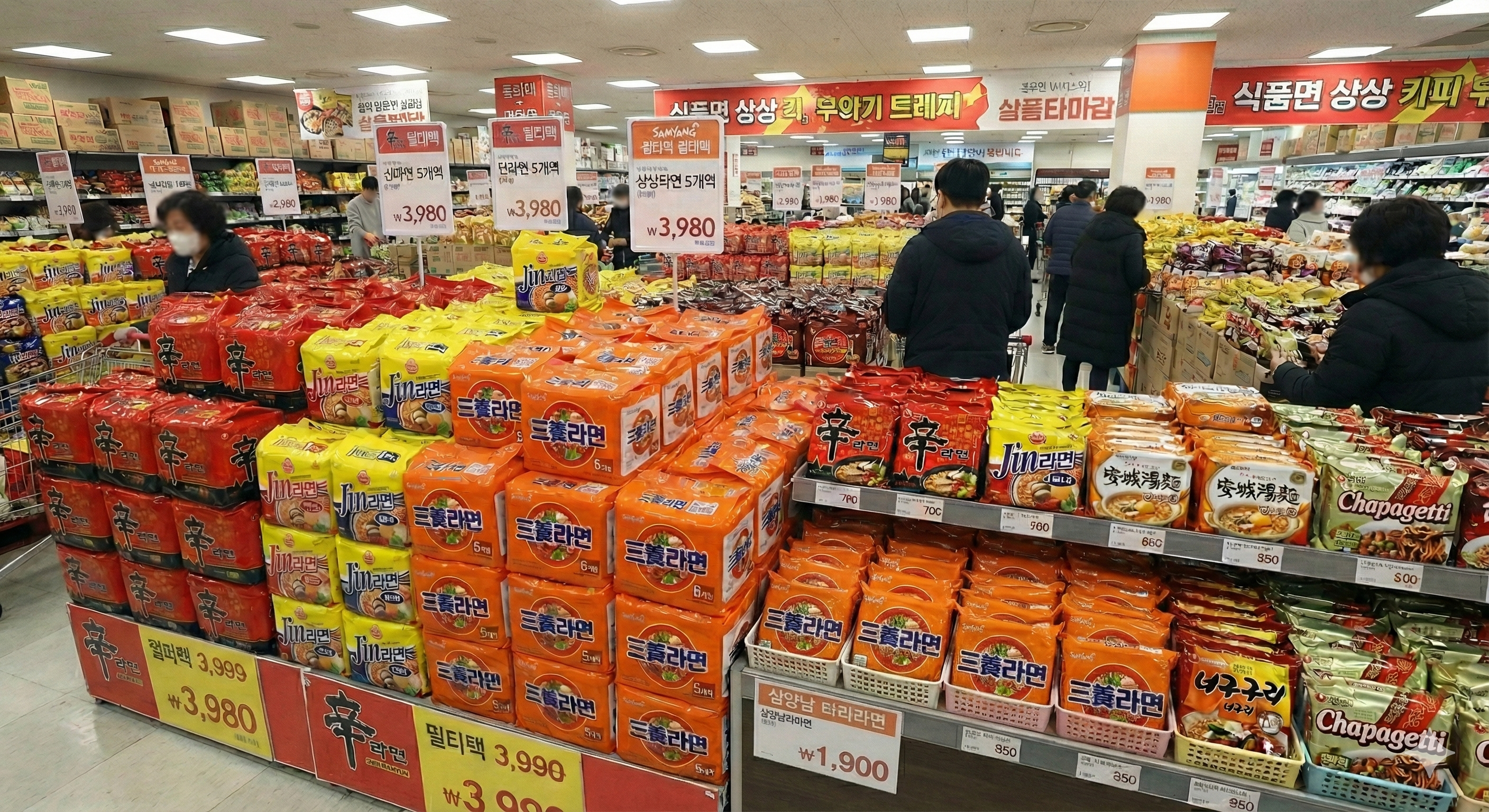}
\caption{AI-generated supermarket ramen-shelf image used as contextual input in the image and image+text conditions.}
\label{fig:ramen_shelf}
\end{figure}

\begin{quote}
\small
\textbf{Example User Prompt (excerpt)}

This is scenario 5 of 12.
Assume that you have already responded to the previous four scenarios and answer consistently with your previous responses.

Your baseline brand preferences are:

Shin Ramyun: 50\%

Samyang Ramen: 30\%

Jin Ramen: 20\%

Imagine that you are shopping for groceries for the coming week and encounter the following ramen products:

Shin Ramyun (Nongshim) --- 610 KRW

Samyang Ramen (Samyang Foods) --- 590 KRW

Jin Ramen (Otoki) --- 570 KRW

Promotion:

Buy two packs and receive one additional pack for free.

Q1. Which product would you purchase?

1) Shin Ramyun

2) Samyang Ramen

3) Jin Ramen

4) No purchase

Q2. How many packs would you purchase?

Please return only two numeric responses corresponding to the answers for Q1 and Q2.
\end{quote}

\subsection{Persona Representation}
\label{app:persona_representation}
Respondent information was represented using either a natural-language text profile or a structured JSON object. Both formats contained identical information, including demographic attributes, household characteristics, weekly ramen consumption, and attitudinal measures related to product involvement and purchasing tendencies. Baseline brand preference allocations were provided separately in the prompt and were therefore not included in the persona representation itself.

The text-persona condition presented respondent attributes as a short natural-language description, whereas the JSON condition represented the same information as key--value pairs. Aside from the representation format, no information differed between the two conditions. Examples of both persona formats are shown below.

\begin{quote}
\small
\textbf{Example Text Persona}

Male, 28 years old, single, monthly income of 0 KRW, purchases 5 packs of ramen per week. Shows moderately high interest in ramen products and perceives ramen as somewhat relevant to daily life. Slightly dislikes adventurous behavior, moderately prefers purchase certainty, and is moderately likely to purchase food products directly.
\end{quote}

\begin{quote}
\small
\textbf{Example JSON Persona}

\begin{verbatim}
{
  "gender": "male",
  "occupation": "student",
  "marital_status": "single",
  "age": 28,
  "household_size": 3,
  "monthly_income": 0,
  "weekly_ramen_purchase": 5,
  "ramen_interest": "somewhat agree",
  "ramen_life_relevance": "somewhat agree",
  "adventurousness": "somewhat disagree",
  "purchase_certainty": "somewhat agree",
  "direct_food_purchase": "somewhat agree"
}
\end{verbatim}
\end{quote}

\subsection{Reasoning Prompt Variants}
\label{app:reasoning_variants}
We evaluated three reasoning-prompting conditions: none, trace-free, and trace-guided. In the none condition, models directly answered the survey questions without generating an intermediate reasoning trace. In the two reasoning conditions, the model first generated a respondent-level background trace before answering any promotional condition. This trace was generated from the system prompt containing the persona and baseline brand preferences, and was then inserted into the user prompt for all 12 promotional scenarios for that respondent.

The trace-free condition asked the model to infer broad background assumptions about the respondent and the 2010 Korean supermarket context, without specifying which aspects had to be included. The model was explicitly instructed not to make a final brand choice or quantity decision during this stage. The trace-guided condition used a more structured instruction, asking the model to distinguish between general assumptions about ramen purchasing in a Korean supermarket context and respondent-specific assumptions inferred from the persona. It also prohibited unsupported personal events, specific past brand experiences, final brand choices, promotion-specific conclusions, and purchase quantities.

In both reasoning conditions, the generated trace was provided as background context before the final survey questions. The final decision prompt retained the same output constraint as the no-reasoning condition: models had to return only two numeric responses corresponding to brand choice and purchase quantity.

\begin{quote}
\small
\textbf{Trace-free instruction (excerpt)}

Generate as many plausible background assumptions as possible about the consumer, based on the persona information, baseline ramen preferences, and the 2010 Korean supermarket context. Write these as assumptions rather than facts. Do not include any conclusion about which ramen brand the consumer would choose or how many packs they would purchase.
\end{quote}

\begin{quote}
\small
\textbf{Trace-guided instruction (excerpt)}

Generate background assumptions for later purchase judgment. First, infer general consumer perceptions of ramen and ramen promotions in a 2010 Korean supermarket context. Second, infer respondent-specific characteristics from the persona, such as economic autonomy, household shopping purpose, direct food purchasing tendency, ramen involvement, adventurousness, and preference for purchase certainty. Do not include final brand choices, purchase quantities, or promotion-specific conclusions.
\end{quote}

\subsection{Model and Decoding Settings}
\label{app:model_details}
The experiments were conducted using GPT-5.2, GPT-5-nano, GPT-4.1-nano, Gemini-2.5-Pro, Qwen2.5-VL-72B-Instruct, Qwen3-VL-30B-A3B-Instruct, Qwen3-VL-8B-Instruct, Llama-3.2-11B-Vision-Instruct, and Llama3-LLaVA-
Next-8B-HF. OpenAI and Gemini models were accessed through their respective APIs. Open-source models were downloaded from Hugging Face and executed locally using the Transformers framework.

All open-source models were run on a workstation equipped with an NVIDIA RTX PRO 6000 Blackwell Workstation Edition GPU. For open-source models, generation used temperature sampling with temperature set to 1.0. API-based models used the providers' default decoding settings. No additional top-$p$, frequency-penalty, presence-penalty, or seed values were manually specified.

For survey-response generation, the maximum output length was limited to 32 tokens because models were required to return only a brand-choice response and a purchase quantity. For reasoning conditions, intermediate reasoning traces were generated separately with a maximum output length of 2048 tokens.

For image-based conditions, all models received the same supermarket ramen-shelf image together with the textual prompt. No additional OCR or manual image preprocessing was applied.

\subsection{Sampling Procedure}
For each model and input configuration, responses were generated for all 52 respondent profiles across the 12 promotional conditions. Each respondent-condition pair was sampled five times, yielding 3,120 generated responses per model and input configuration ($52 \times 12 \times 5$). The five repeated runs were aggregated to construct response distributions for purchase incidence, brand choice, and purchase quantity.

For reasoning conditions, a respondent-level reasoning trace was generated once per respondent and run. The same trace was then inserted into the prompts for all 12 promotional conditions for that respondent within that run. This procedure preserves a consistent respondent-level background across the full set of promotional scenarios while still allowing repeated sampling across runs.

No fixed random seed was specified. API or inference failures were retried up to two times using exponential backoff. Responses were then aggregated by model, input configuration, and promotional condition for the evaluation metrics described in Section~\ref{sec:evaluation_metrics}.

\section{Input Modality Analysis}
\label{app:modality_results}
We report the effect of input modality on replication fidelity, which produced the smallest effect among the three input dimensions analyzed in Section~\ref{sec:input_conditions}. We compare three modalities: text-only, in which promotional information is given as a natural-language description; image-only, in which it is rendered as a product display of the promotional offer; and image+text, which combines both. As in Section~\ref{sec:input_conditions}, results are averaged across all models, promotional conditions, and response types.
Table \ref{tab:input_modality} reports the results. Both image-based conditions slightly outperform text-only on both metrics. The image+text condition achieves the highest mean Pearson correlation (0.750), the image-only condition the lowest mean JSD (0.272), and the text-only condition the weakest alignment on both metrics (0.707 Pearson, 0.285 JSD). The differences are small in magnitude: about 0.04 in Pearson correlation and 0.01 in JSD between text-only and image+text, but the direction of the effect is consistent across both alignment levels.
These results suggest that visual presentation of promotional information contributes incremental rather than transformative signal. The near-identical performance of the image-only and image+text conditions further indicates that adding textual description to image input does little beyond what image input alone provides, implying that the two modalities convey largely overlapping information about the promotional offer in this task.

\begin{table}[t]
\centering
\footnotesize
\setlength{\tabcolsep}{3pt}
\renewcommand{\arraystretch}{1.2}

\begin{tabular}{
>{\raggedright\arraybackslash}m{2.0cm}
>{\centering\arraybackslash}m{1.7cm}
>{\centering\arraybackslash}m{1.7cm}
}
\hline
\centering\textbf{Input modality}
& \textbf{Mean Pearson $r$}
& \textbf{Mean JSD} \\
\hline

Image & 0.739 & 0.272 \\
Image+text & 0.750 & 0.274 \\
Text & 0.707 & 0.285 \\

\hline
\end{tabular}

\caption{
Average pattern and distributional alignment across input modalities.
}

\label{tab:input_modality}
\end{table}

\section{Bootstrap Confidence Intervals}
\label{app:bootstrap_ci}

To assess the stability of the distributional alignment metrics,
we computed 95\% bootstrap confidence intervals by resampling
the 12 promotional conditions with replacement 10,000 times.
Tables~\ref{tab:binary_ci}, \ref{tab:brand_ci}, and
\ref{tab:quantity_ci} report the resulting confidence intervals
for binary purchase incidence, brand choice, and purchase
quantity, respectively.

\begin{table}[t]
\centering
\footnotesize

\begin{tabular}{
>{\raggedright\arraybackslash}m{2.8cm}
>{\centering\arraybackslash}m{1.4cm}
>{\centering\arraybackslash}m{2.2cm}
}
\toprule
\textbf{Model} & \textbf{Mean JSD} & \textbf{95\% CI} \\
\midrule
Gemini-2.5-Pro & 0.003 & [0.002, 0.005] \\
Qwen3-VL-30B-A3B-Instruct & 0.013 & [0.008, 0.019] \\
Qwen2.5-VL-7B-Instruct & 0.018 & [0.007, 0.035] \\
GPT-5-nano & 0.021 & [0.017, 0.026] \\
Qwen2.5-VL-72B-Instruct & 0.021 & [0.014, 0.028] \\
GPT-5.2 & 0.024 & [0.020, 0.029] \\
Llama-3.2-11B-Vision-Instruct & 0.025 & [0.020, 0.030] \\
Qwen3-VL-8B-Instruct & 0.025 & [0.020, 0.030] \\
GPT-4.1-nano & 0.025 & [0.020, 0.030] \\
Llama3-LLaVA-Next-8B-HF & 0.025 & [0.020, 0.030] \\
\bottomrule
\end{tabular}

\caption{
95\% bootstrap confidence intervals for binary purchase incidence
distributional alignment (JSD).
}
\label{tab:binary_ci}
\end{table}

\begin{table}[t]
\centering
\footnotesize

\begin{tabular}{
>{\raggedright\arraybackslash}m{2.8cm}
>{\centering\arraybackslash}m{1.4cm}
>{\centering\arraybackslash}m{2.2cm}
}
\toprule
\textbf{Model} & \textbf{Mean JSD} & \textbf{95\% CI} \\
\midrule
Qwen2.5-VL-72B-Instruct & 0.014 & [0.007, 0.022] \\
Llama-3.2-11B-Vision-Instruct & 0.030 & [0.019, 0.041] \\
Qwen3-VL-8B-Instruct & 0.031 & [0.020, 0.044] \\
Qwen3-VL-30B-A3B-Instruct & 0.040 & [0.024, 0.058] \\
GPT-5-nano & 0.056 & [0.023, 0.097] \\
GPT-5.2 & 0.075 & [0.029, 0.134] \\
Gemini-2.5-Pro & 0.081 & [0.046, 0.122] \\
GPT-4.1-nano & 0.103 & [0.068, 0.141] \\
Llama3-LLaVA-Next-8B-HF & 0.262 & [0.181, 0.342] \\
Qwen2.5-VL-7B-Instruct & 0.363 & [0.280, 0.456] \\
\bottomrule
\end{tabular}

\caption{
95\% bootstrap confidence intervals for brand choice
distributional alignment (JSD).
}
\label{tab:brand_ci}
\end{table}

\begin{table}[t]
\centering
\footnotesize

\begin{tabular}{
>{\raggedright\arraybackslash}m{2.8cm}
>{\centering\arraybackslash}m{1.3cm}
>{\centering\arraybackslash}m{2.0cm}
}
\toprule

\textbf{Model}
&
\makecell{\textbf{Mean}\\\textbf{Wasserstein}}
&
\textbf{95\% CI}
\\

\midrule
Gemini-2.5-Pro & 1.467 & [1.251, 1.698] \\
Llama-3.2-11B-Vision-Instruct & 1.669 & [1.373, 1.980] \\
Qwen3-VL-30B-A3B-Instruct & 1.751 & [1.488, 2.091] \\
Qwen3-VL-8B-Instruct & 1.810 & [1.535, 2.116] \\
GPT-5-nano & 1.819 & [1.516, 2.187] \\
GPT-5.2 & 1.968 & [1.702, 2.293] \\
Qwen2.5-VL-72B-Instruct & 2.064 & [1.776, 2.417] \\
Qwen2.5-VL-7B-Instruct & 2.503 & [2.158, 2.888] \\
GPT-4.1-nano & 2.977 & [2.553, 3.464] \\
Llama3-LLaVA-Next-8B-HF & 3.155 & [2.826, 3.530] \\
\bottomrule
\end{tabular}

\caption{
95\% bootstrap confidence intervals for purchase quantity
distributional alignment (Wasserstein distance).
}
\label{tab:quantity_ci}
\end{table}

\end{document}